\documentclass{article}

\usepackage[final]{neurips_2021}

\setcitestyle{square,numbers,comma}
\usepackage[utf8]{inputenc} 
\usepackage[T1]{fontenc}    
\usepackage{hyperref}       
\usepackage{url}            
\usepackage{booktabs}       
\usepackage{amsfonts}       
\usepackage{graphicx}
\usepackage{nicefrac}       
\usepackage{microtype}      
\usepackage{xcolor}         
\usepackage{booktabs}       
\usepackage{subcaption}
\usepackage{cleveref}

\title{
Unsupervised Change Detection of \\Extreme Events Using ML On-Board\\
}
\author{%
  Vít Růžička$^{1}$\thanks{Correspondence to: \href{mailto:vit.ruzicka@cs.ox.ac.uk}{vit.ruzicka@cs.ox.ac.uk}}, Anna Vaughan$^{2}$, Daniele De Martini$^{1}$, James Fulton$^{3}$, \\
  
  \textbf{Valentina Salvatelli$^{4}$,
  Chris Bridges$^{5}$,
  Gonzalo Mateo-Garcia$^{6}$,
  Valentina Zantedeschi$^{7}$} \\
  ${}^1$University of Oxford, \hfill ${}^2$University of Cambridge,\hfill ${}^3$University of Edinburgh,\hfill ${}^4$Microsoft, \\
  ${}^5$University of Surrey,\hfill ${}^6$University of Valencia, \hfill${}^7$INRIA, University College London \\

}

\newcommand{\ravaen}{\textbf{RaV{\AE}n}}

\begin{document}

\maketitle

\begin{abstract}
In this paper, we introduce \ravaen, a lightweight, unsupervised approach for change detection in satellite data based on Variational Auto-Encoders (VAEs) with the specific purpose of on-board deployment.
Applications such as disaster management enormously benefit from the rapid availability of satellite observations.
Traditionally, data analysis is performed on the ground after all data is transferred -- downlinked -- to a ground station. Constraint on the downlink capabilities therefore affects any downstream application.
In contrast, \ravaen~pre-processes the sampled data directly on the satellite and flags changed areas to prioritise for downlink, shortening the response time.
We verified the efficacy of our system on a dataset composed of time series of catastrophic events -- which we plan to release alongside this publication -- demonstrating that \ravaen~outperforms pixel-wise baselines. Finally we tested our approach on resource-limited hardware for assessing computational and memory limitations. 
\end{abstract}

\section{Introduction}

Satellite observations of the Earth's surface provide a vital data source for diverse environmental applications, including disaster management, landcover change detection, and ecological monitoring.
Currently, sensors on satellites collect and downlink all data points for further processing on the ground; yet, limitations in downlink capacity and speed result in delayed data availability and inefficient use of ground stations.
These limitations affect time-sensitive applications such as disaster management where data are required with as low latency as possible to inform decision making in real time. 
This problem is set to worsen as the sensing resolution and number of satellites in orbit increase together with further restrictions on radio-frequency spectrum sharing licensing. 
 
A solution is to apply processing on-board to identify the most useful data for a particular scenario, and prioritise this for rapid downlink. There has been considerable recent interest in using machine learning on-board for this processing. Existing work has focused on the deployment of supervised classifiers for applications such as identifying clouds~\cite{giuffrida_cloudscout_2020} or floods~\cite{worldfloods_2021}.
Supervised learning has a significant drawback: only events of a particular type determined at training time will be flagged and prioritised for downlink, with no generalisation to new event types, imager specifications, and lighting or local features.

In this work we propose a new fully unsupervised novelty-detection model suitable for deployment on remote sensing platforms.
We use a \textit{Variational AutoEncoder} (VAE)~\citep{kingma2013autoencoding} to generate a latent representation of incoming sensor data over a particular region.
A novelty score is assigned to this data using the distance in the latent space between representations from consecutive passes. This offers a substantial advantage over existing supervised methods as any change between passes can be detected on-board at inference time, regardless of the availability of training data.

We evaluate the performance of this model detecting changes in land-surface observations from the Sentinel-2 Multispectral Instrument on time series of natural disaster events where obtaining data with low-latency is of importance, including floods, landslides, wildfires and hurricanes.
The unsupervised novelty detection model is demonstrated to assign higher novelty scores to regions of known change and outperforms computer-vision image differencing baselines. We further demonstrate via experiments on constrained Xilinx Pynq hardware that this model is suitable for deployment on a remote sensing platform. 

\subsection{Application Context}

Our proposed method \ravaen~facilitates intelligent decision making on board of a satellite for rapid disaster response.
With this application in mind it has been tested on constrained hardware to similar to that available on remote sensing platforms. 
Using this system we can detect extreme events and rapidly prioritize data over the affected regions for downlink. Vital data will therefore be available with lower latency to inform decision making during rapidly evolving events. In the rest of this section we frame our work in the context of machine learning literature.

\paragraph{Anomaly detection} The use of VAEs has been explored for unsupervised anomaly detection in \cite{lesnikowski2020unsupervised}, where the model reconstruction error has been used as anomaly score. 
Our approach differs in that, instead of basing our predictions on the reconstruction error of a single input, which has been shown in \cite{merrill2020modified} to be an unreliable indicator in the unsupervised context, we consider a sequence of input images from the same location. Our problem is better framed as change detection.

\paragraph{Change detection} The need for annotations of supervised change detection techniques, such as siamese networks in \cite{Caye_Daudt_2018}, can be reduced using active learning approaches as demonstrated in~\cite{ruzicka2020deep}, but then it still lacks in terms of generality. 
The main challenge of unsupervised change detection is being able to distinguish changes of interest from spurious change due to noise. 
Many existing approaches~\citep{celik2009unsupervised,CelikC10,ChengLCZ13} achieve this by combining dimensionality reduction techniques, such as Principal Component Analysis, and clustering, such as $k$-means, to detect only relevant change between images of consecutive passes.
Approaches based on neural networks (see \cite{de2019unsupervised} for a review) rely instead on supervised auxiliary tasks, such as semantic segmentation, to extract informative features that are then used to detect change in a time series.
Our method leverages neural networks without requiring supervision at any stage. 

\section{Data}
\label{sec:data}

\begin{figure}
    \centering
    \begin{subfigure}{0.49\textwidth}
    \includegraphics[width=\textwidth]{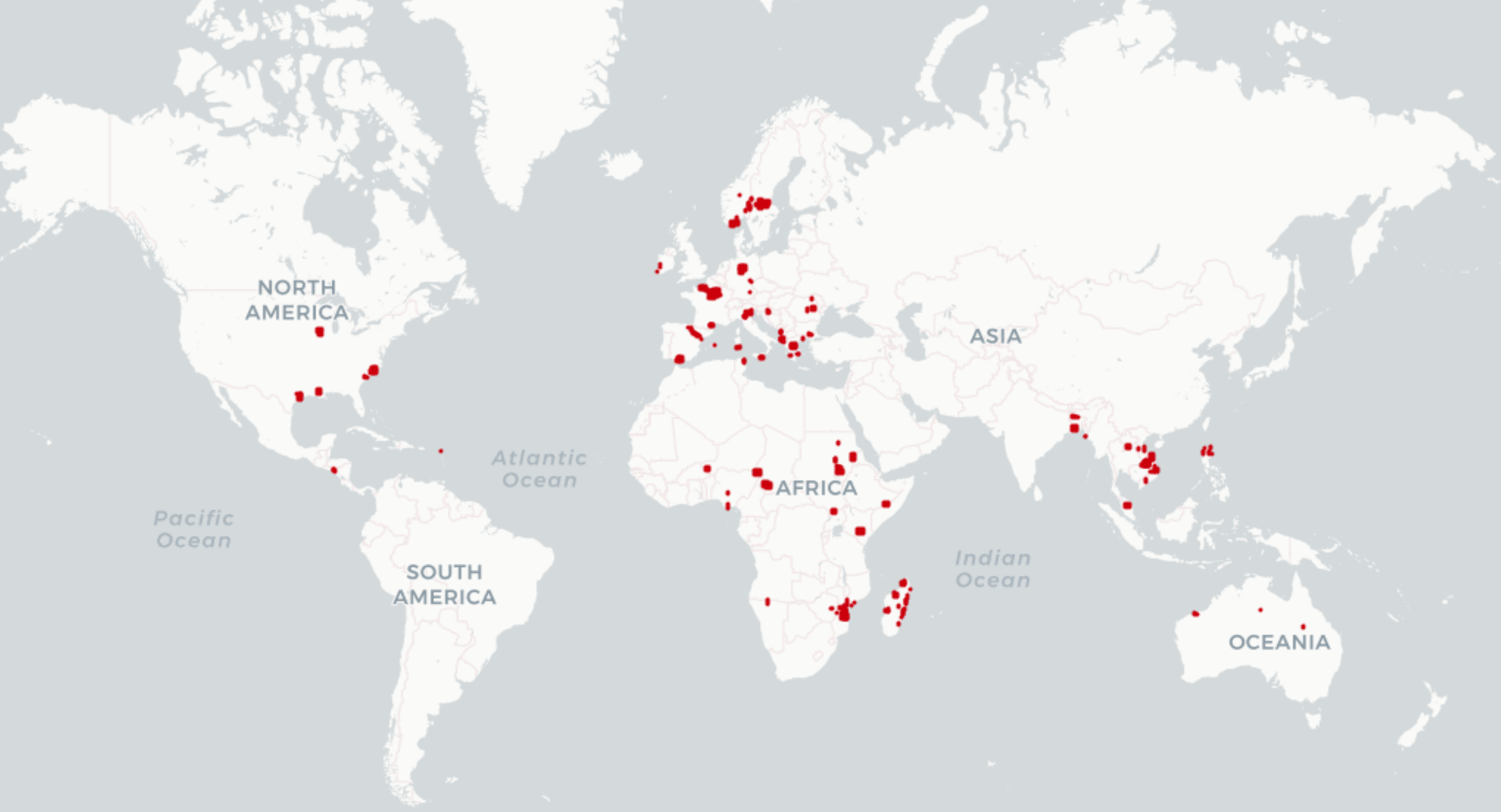}
    \caption{\label{fig:worldfloods}}
    \end{subfigure}
    \begin{subfigure}{0.49\textwidth}
    \includegraphics[width=\textwidth]{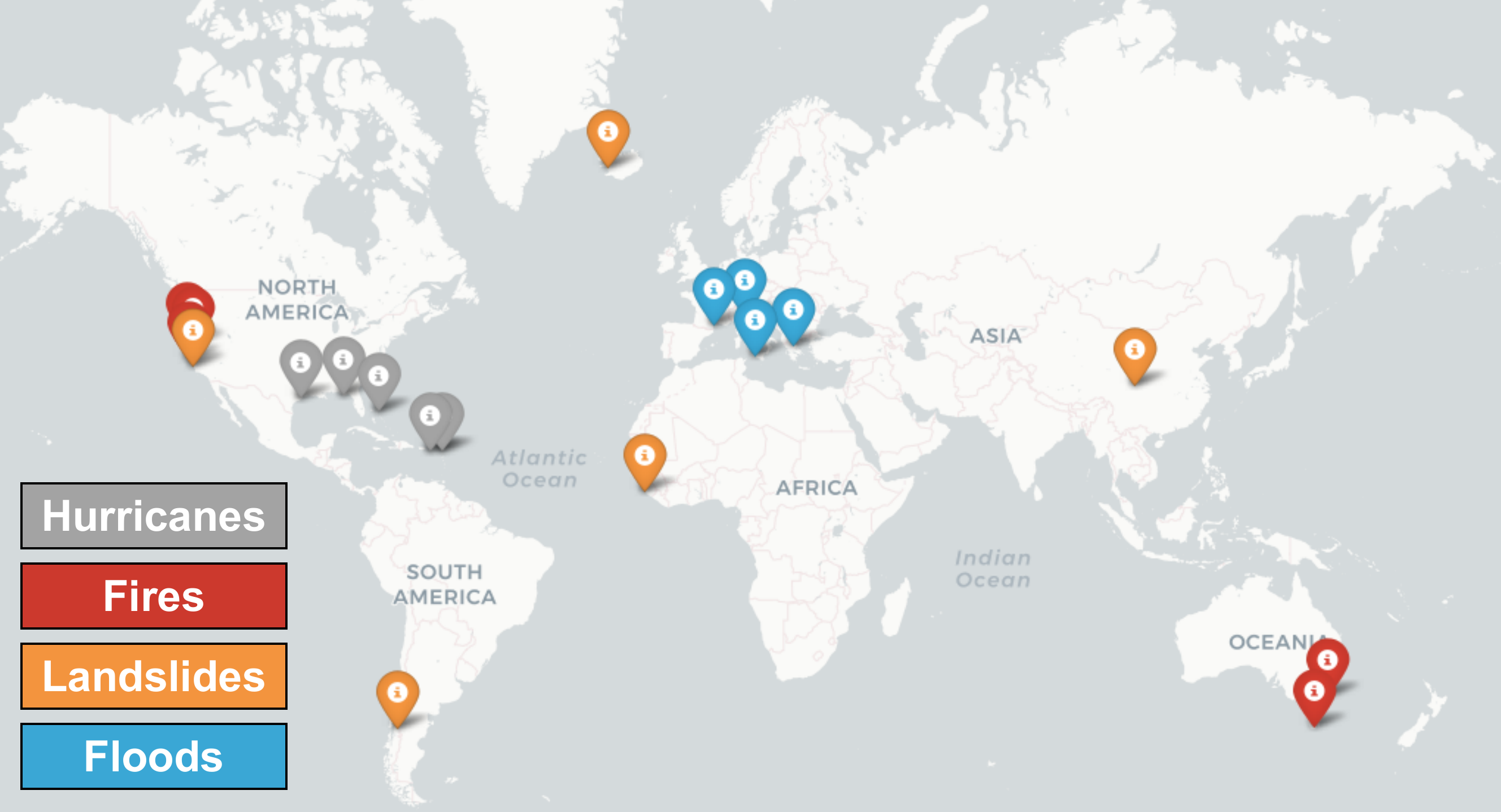}
    \caption{\label{fig:testset}}
    \end{subfigure}
    \caption{Locations used for training (a) and validation (b) images.\label{fig:my_label}}
\end{figure}

As part of this study, we compile a new dataset to evaluate the proposed unsupervised change detection models. 
Images are taken from the Sentinel-2 multi-spectral imager (MSI) instrument\footnote{\url{https://sentinels.copernicus.eu/web/sentinel/user-guides/sentinel-2-msi/overview}} and include the ten highest resolution channels with all channels interpolated to the highest resolution of 10m. Training data are taken from the \textit{WorldFloods} dataset of \cite{worldfloods_2021} locations (\Cref{fig:worldfloods}), with a total of 233 scenes and a time series of five images per scene. 

\begin{figure}
    \centering
    \includegraphics[width=0.8\textwidth]{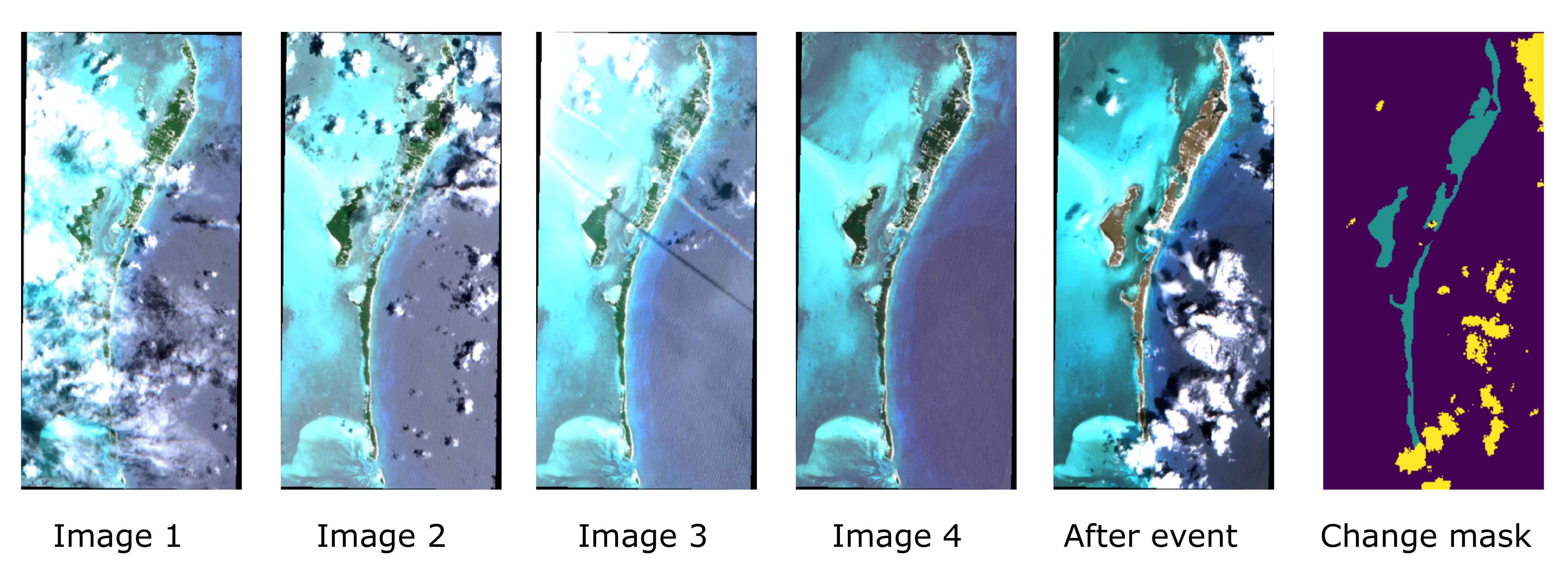}
    \caption{Example of validation sample -- in this case, a hurricane event -- and its corresponding ground-truth mask (which contains labels of change and clouds).\label{fig:hurricane}}
\end{figure}

The validation set, consists of Sentinel-2 time series for four classes of disasters: hurricanes, fire burn scars, landslides and floods (\Cref{fig:testset}).
We identified events in each of these classes through an extensive search of Sentinel-2 records aided by the Copernicus EMS system\footnote{\url{https://emergency.copernicus.eu/mapping/list-of-activations-rapid}}.
Each event in the validation set consists of a time series of five images where the first four images are taken before the disaster while the fifth image is taken afterwards.
To mitigate the effects of cloud cover, we discarded validation images with greater than 20\% cloud cover.
Events are only included where all images are within 180 days before and 90 days after the event.
For each event a change mask was hand annotated to mark corresponding to the difference between the final two images in the time series, for example \Cref{fig:hurricane}.
We emphasize that these labels are used for evaluation only. The distribution of data in the validation dataset is further detailed in Appendix \ref{sec:dataset-description}.

\section{Methodology}
\label{sec:methodology}


\paragraph{Preprocessing} 
Tiles of 32x32 pixels are extracted from the Sentinel-2 scenes described above and used as inputs to the considered models. 
These are further normalized by applying a log transform and scaling to constrain them to the $[0,1]$ interval. 


\paragraph{Model} We use a VAE (illustrated in Figure \ref{fig:architecture} of the appendix) to learn an meaningful embedding space for change detection. The VAE is chosen as it guarantees a meaningful distance metric on the embedding space.
The model is trained to reconstruct the original input tile from its compressed representation distribution in the embedding space. 
Once trained we use this model as a feature extractor to encode individual tiles, e.g. $x_t^{a,b}$ and $x_{t-1}^{a,b}$, and compare their latent representations to measure the change between time-steps $t$ and $t-1$. Parameters $a, b$ correspond to the indexed location in the original image.
Comparing compressed representations of images improves robustness to noise and reduces the computational and memory requirements of storing images from previous passes, which is a critical in a constrained environment.
We denote by $\mathcal{N}_t^{a,b} = \mathcal{N}(\mu_t^{a,b}, \sigma_t^{a,b})$ the encoded Gaussian distribution of tile $x_t^{a,b}$ with $\mu_t^{a,b}, \sigma_t^{a,b} \in \mathbb{R}^n$.
In the following, we fix the latent size to $n=128$ after the initial experiments indicated that larger latent sizes did not yield improved results.

\paragraph{Change detection novelty score} We compute a change score for tiles $x_{t-1}^{a,b}, x_{t}^{a,b}$ using either the KL divergence between encoded distributions $\mathcal{N}_{t-1}^{a,b}$ and $\mathcal{N}_t^{a,b}$, or the Euclidean or cosine distance between mean encoded vectors $\mu_{t-1}^{a,b}$ and $\mu_t^{a,b}$, that we generally note as $\mathrm{D}_{\mathrm{VAE}}(x_{t-1}^{a,b}, x_t^{a,b})$.
We extend the methodology to use a longer time series of $k$ past images by retaining the minimal distance as:
\begin{equation}
\label{eqn:score_latentspace}
\mathrm{S}_{\mathrm{VAE}}(x_t^{a,b}) = \min_{i = 1 ... k} \mathrm{D}_{\mathrm{VAE}}(x_{t-i}^{a,b}, x_t^{a,b})
\end{equation}

To compare the performance of this approach to simpler on-board processing methods that does not make use of machine learning, we compare our method to a baseline which compares tiles directly in the input space using the Euclidean or the cosine distance and after applying the same data pre-processing as for the VAE.

\begin{figure}[t]
	\centering
	\includegraphics[width=0.9\textwidth]{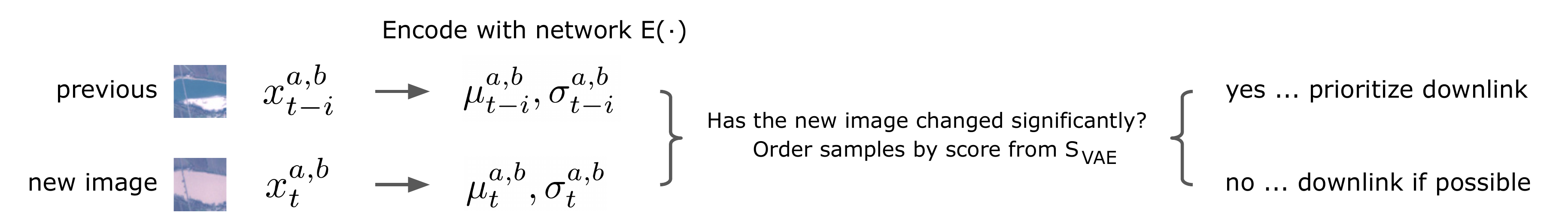}
	\caption{Example of \ravaen,~the suggested decision making process on board of the satellite (for a time window of $k=1$).}
	\label{fig:downlink-decision}
\end{figure}

The proposed workflow outlined in this section is summarised in Figure \ref{fig:downlink-decision}.

\section{Experimental Setting and Reproducibility}

We use different environments for training the VAE and for inference.
For development (training and validation of the model), we use a \texttt{n1-standard-16} instance on Google Cloud Platform with two NVIDIA Tesla V100 GPUs.
In addition we measure the performance of the models on the Xilinx Pynq FPGA board with limited compute power, 650MHz ARM Cortex-A9 CPU and 512MB RAM which emulates the resources available on a typical small satellite. We report the metrics for the methodology deployed on real hardware in Appendix \ref{sec:timings}.

\paragraph{Model architecture design choices}
To minimize the size of the model and maximize efficiency on constrained devices we conducted a parametric search over both the number of layers and number of units per layer in both the encoder network $E$ and the decoder network $D$. 
More precisely we tested three different model architecture configurations (\textit{small}, \textit{medium} and \textit{large}). Details and results of this experiment are included in Appendix \ref{sec:timings}, where we show their maintained performance while reaching fast processing speeds of only $\sim$2 seconds on the Xilinx Pynq FPGA board.
In the following, we report results for the \textit{large} VAE configuration.

\section{Results}
\label{sec:results}

\begin{figure}[t]
	\centering
	\includegraphics[width=0.9\textwidth]{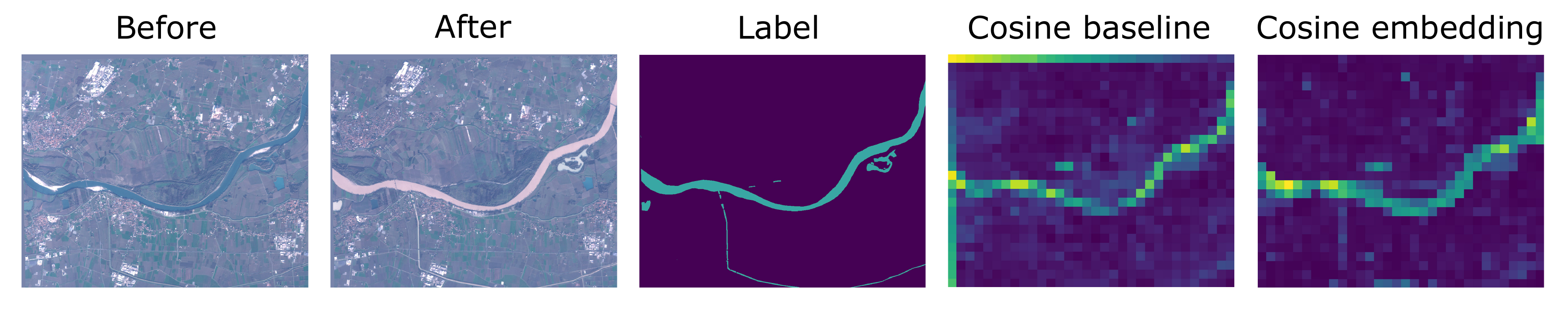}
	\caption{Comparison of the change detected using the baseline and the \textit{large} VAE method on an example of a flooding river. Two images immediately before and immediately after a change are shown, along with the human labels of change and the calculated change scores. Both methods used a history of $k=3$ frames.}
	\label{fig:river-example}
\vspace{-2mm}

\end{figure}

\begin{figure}[t]
	\centering
	\includegraphics[width=0.9\textwidth]{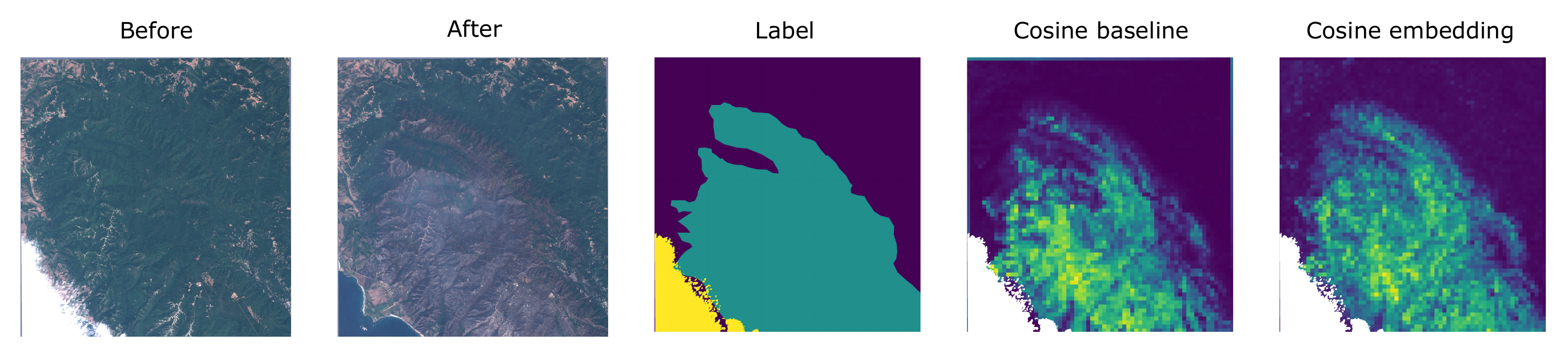}
	\caption{Additional comparison of the change detected using the baseline and the \textit{large} VAE method on an example of a fire disaster. Both methods used a history of $k=3$ frames. The cosine baseline prediction seems to more closely copy the details present in the image, making it susceptible to small, noisy variations between the two images.}
	\label{fig:fire-example}
\end{figure}

\Cref{fig:river-example} shows a qualitative comparison between the VAE model developed in this study and the image differencing baseline.
The \textit{before} image shows a river that floods and therefore changes colour in the \textit{after} image.
The labels and the change scores from our \textit{large} VAE and baseline methods are shown in the bottom row of the figure. 
In this case, the scores were calculated using a history of $k=3$ frames, although only the most recent \textit{before} frame is shown for brevity.
In this example, our method -- the cosine embedding -- produces a change map that is crisper than the cosine baseline; for instance, the small flooded canal can be seen in the cosine embedding image but not in the baseline.
In a similar fashion, \Cref{fig:fire-example} shows a qualitative comparison in the case of a burnt-area detection.

The change-score maps, like those in \Cref{fig:river-example,fig:fire-example}, were produced for every image in the evaluation set.
We use these maps and our labels to calculate the area under the precision-recall curve.
We produce the curve tile-wise, so that each individual tile across each image is treated as a positive or negative example of change, rather than treating the full image as one example.
This means our quality metric is sensitive to the fact that our evaluation images are not equal; they have different number of tiles and different ratios of positive pixels (as reported in \Cref{tab:dataset-eval-summary}).
We also ignore tiles that have clouds in the \textit{after} image or in the most recent image before the event.
We produce a precision-recall curve for each of the four different event types in our evaluation set.

\Cref{tab:AUPR-metric-compare} reports the results of our change detection experiments for all disaster types.
We found that cosine distance, applied on the input space or on the embeddings, generally provides the best scores.
Surprisingly KL-divergence is the lowest-performing metric, and is beaten by both cosine and Euclidean embedding scores in all events, even though these methods do not use the variance values calculated by the VAE. 
Metrics based on the VAE embedding outperforms the baseline on floods, hurricanes and fires, and reaches similar performance on landslides.

\Cref{tab:AUPR-history-compare} shows the effects of including a longer frame history.
When three previous images are provided instead of just one, both the embedding and baseline perform better except for the case of landslide dataset where the cosine baseline with memory 1 beats memory 3 with a small margin.
The table also shows that our method of detecting significant change based on the embedding space outperforms the baseline in every dataset when $k=3$. 

\begin{table}[!htb]
    \caption{Area under the precision-recall curve for baseline and VAE methods with time window $k=1$ (averaged over 5 runs).}
    \label{tab:AUPR-metric-compare}
    \centering
    \begin{tabular}{@{}lllll@{}}
    \toprule
    \textbf{Detection method} & \multicolumn{4}{c}{\textbf{Dataset}}                                       \\
                     &    Landslides     & Floods          & Hurricanes     & Fires          \\
    Cosine baseline      & \textbf{0.629}             & 0.378                      & 0.513                       & 0.818 \\
    Euclidean baseline        & 0.267                      & 0.326                      & 0.351                       & 0.770 \\
    Cosine embedding     & 0.599 $\pm$ 0.012          & \textbf{0.448} $\pm$ 0.011 & \textbf{0.676} $\pm$ 0.014  & \textbf{0.833} $\pm$ 0.008 \\
    Euclidean embedding       & 0.266 $\pm$ 0.004          & \textbf{0.450} $\pm$ 0.007 & 0.478 $\pm$ 0.019           & 0.800 $\pm$ 0.011 \\
    KL-Divergence        & 0.258 $\pm$ 0.022          & 0.247 $\pm$ 0.018          & 0.301 $\pm$ 0.035           & 0.731 $\pm$ 0.016 \\
    \bottomrule
    \end{tabular}
\vspace{-2mm}
\end{table}

\begin{table}[!htb]
    \caption{Area under the precision-recall curve for the best performing metrics from table \ref{tab:AUPR-metric-compare} with and without an extended history $k$ (averaged over 5 runs).}
    \label{tab:AUPR-history-compare}
    \centering
    \begin{tabular}{@{}lcllll@{}}
    \toprule
    \textbf{Detection method} & \textbf{k} & \multicolumn{4}{c}{\textbf{Dataset}}   \\
            &  & Landslides     & Floods         & Hurricanes     & Fires          \\
    Cosine baseline  & 1      & 0.629          & 0.378          & 0.513          & 0.818          \\
                     & 3      & 0.622          & 0.378          & 0.570          & 0.865          \\
    Cosine embedding & 1      & 0.599 $\pm$ 0.012          & \textbf{0.448} $\pm$ 0.011 & 0.676 $\pm$ 0.014          & 0.833 $\pm$ 0.008          \\
                     & 3      & \textbf{0.759} $\pm$ 0.024 & \textbf{0.443} $\pm$ 0.009 & \textbf{0.726} $\pm$ 0.011 & \textbf{0.913} $\pm$ 0.008 \\
    \bottomrule
    \end{tabular}
\vspace{-2mm}
\end{table}

\section{Conclusion}

In conclusion, we introduce a new method \ravaen~for unsupervised change detection in remote sensing data using a VAE. 
Our method is evaluated on a new dataset of remote sensing images of disasters which we aim to release with our work. 
The proposed model outperforms a classical computer vision baseline in all of the tested disaster classes. This demonstrates that \ravaen~is a robust change detection method and suitable for application in improving data acquisition for disaster response. 

\begin{ack}
This work has been enabled by \href{https://fdleurope.org}{Frontier Development Lab (FDL) Europe}, a public partnership between the European Space Agency (ESA) Phi-Lab (ESRIN) and ESA Mission Operations (ESOC), Trillium Technologies and the University of Oxford; the project has been also supported by Google Cloud, D-Orbit and Planet.
The authors would like to thank all FDL faculty members, Atılım Güneş Baydin and Yarin Gal (University of Oxford), Chedy Raissi (INRIA), Brad Neuberg (Planet) and Nicolas Longépé (ESA ESRIN) for discussions and comments throughout the development of this work.
\end{ack}

\bibliographystyle{unsrtnat}

\bibliography{biblio}

\newpage
\appendix
\section{Appendix}
%
%

\subsection{Model architecture}
\label{sec:model-arch}

\begin{figure}
    \centering
    \includegraphics[width=0.8\textwidth]{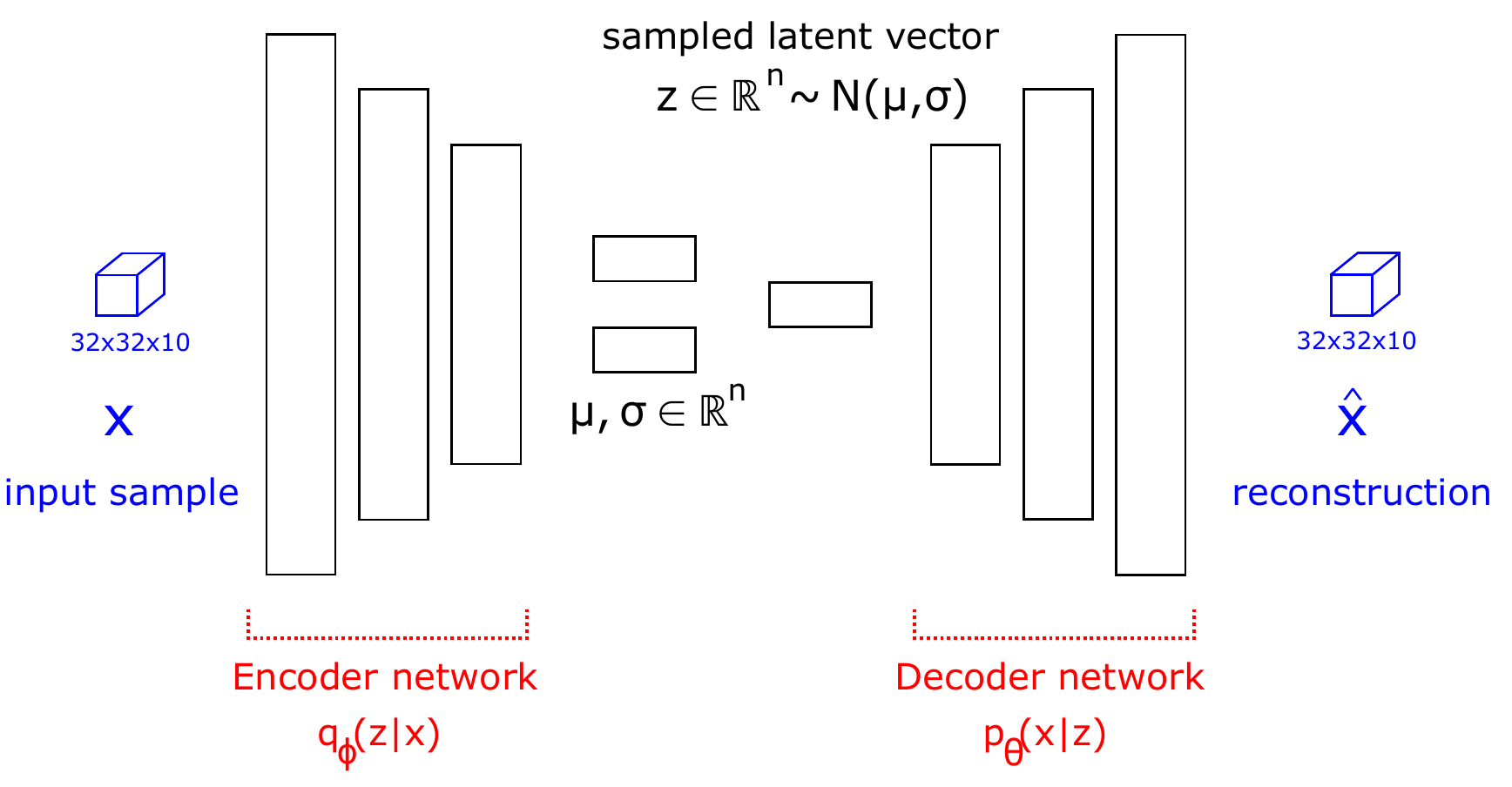}
    \caption{Illustration of the VAE model, $n$ denotes the latent dimension. Number and type of the used layers can be adjusted given the limitations of the used hardware as is explored in more detail in \ref{sec:timings}.}
    \label{fig:architecture}
\end{figure}

The encoder of our VAE was composed of a series of downsampling blocks. Each downsampling block first had a 2D convolutional layer with kernel size 3, stride size 2, and zero padding of 1, such that the dimensions are halved in the spatial domain. Following this layer, the block also had a sequence of extra 2D convolutional layers (the number \textit{extra depth} referred to in table \ref{tab:model-size-summary}). 
Skip connections were used so that the \textit{extra depth} convolutional layers formed a residual block. The network could then easily learn to skip these non-downsampling layers. In the residual block, the number of hidden channels and image size were conserved. Each convolution layer used leaky ReLU activations and batch normalisation. Following a given number of downsampling blocks, the result was flattened and further reduced in dimension using a fully connected layer which output the mean and log variance. The decoder was essentially the encoder in reverse. The upsampling method used was nearest neighbour upsampling followed by a single convolution. This method was preferred over transpose convolution to avoid checkerboard artefacts \cite{odena2016deconvolution, wang2020cnn}.

The main model presented in this paper is denoted as \textit{large} on Table \ref{tab:model-size-summary}, it used 3 downsampling blocks with 32, 64, 128 channels on each successively smaller scale. The network was further squeezed via the fully connected layer to a latent dimension of 128. We used extra depth value of 2, so after each downscale convolution there was a residual block of 2 additional convolutional layers.

\subsection{Model timings}
\label{sec:timings}

\begin{table}[!htb]
    \caption{Area under the precision recall curve and timings for different sizes of model (averaged over 5 runs). The AUPRC results are for the cosine similarity of the embedding with a history of 3 frames.}
    \label{tab:accuracy-and-timings}
    \centering
    \begin{tabular}{@{}lllllrrrr@{}}
    \toprule
                        & \multicolumn{4}{c}{\textbf{Dataset}}  \\
                        
                        & Landslides                  & Floods                      & Hurricanes                  & Fires \\
    Small model         & \textbf{0.748} $\pm$ 0.014  & \textbf{0.445} $\pm$ 0.014  & \textbf{0.748} $\pm$ 0.002  & 0.907 $\pm$ 0.002  \\
    Medium model        & \textbf{0.758} $\pm$ 0.007  & 0.428 $\pm$ 0.004           & \textbf{0.738} $\pm$ 0.018  & \textbf{0.912} $\pm$ 0.003  \\
    Large model         & \textbf{0.759} $\pm$ 0.024  & \textbf{0.443} $\pm$ 0.009  & 0.726 $\pm$ 0.011           & \textbf{0.913} $\pm$ 0.001  \\
    \bottomrule
    \end{tabular}
\end{table}

It is strongly desired that this change detection method could be run onboard a satellite, so that images can be filtered or prioritised before they reach the ground. Therefore the time it takes to run change detection is of the utmost importance. We need to design our models so that the low power hardware, available onboard a mission, can keep up with the incoming stream of data. We therefore tested the accuracy of our model with different numbers of channels, and different numbers of layers.

Table \ref{tab:accuracy-and-timings} shows the accuracy of a few variations of model size and the time it took to process a 574$\times$509 px image (approx. 5km$\times$5km at Sentinel-2 10m resolution) whilst running on the CPU of a  Xilinx PYNQ. 
We see the the results of all tested models are comparable and that it is reasonable to aim for the smallest model, which takes only 2.06 seconds to process the patch. Running onboard the PYNQ means that there is considerable potential to speed up this runtime by a large factor by deploying directly on the FPGA module rather than using the board's CPU.

\begin{table}[!htb]
    \caption{Differences in the architecture for different models of table \ref{tab:accuracy-and-timings}. Note that during inference, we only need the encoder network of the VAE model. We also only need to process the newly acquired image to obtain their latent representation, while the latent vectors of the previous image can be loaded.}
    \label{tab:model-size-summary}
    \centering
    \begin{tabular}{@{}llllll@{}}
    \toprule
                        & Total      & Encoder    &  Runtime & Extra depth & Hidden channels   \\ 
                        & parameters & parameters &  (seconds)&           \\                        
                        & (millions) & (millions) &           &   &                \\
    Small model         & 0.443       & 0.285         &   2.06     & 0 & 16, 32, 64     \\
    Medium model        & 0.979       & 0.617         &  4.86     & 0 & 32, 64, 128    \\
    Large model         & 1.500       & 1.005        &  13.98     & 2 & 32, 64, 128    \\
    \bottomrule
    \end{tabular}
\end{table}

\subsection{Dataset description}
\label{sec:dataset-description}

We describe the statistics of the manually annotated validation dataset in Table \ref{tab:dataset-eval-summary}. While each type of the event is represented by similar amount of locations, the affected area varies dramatically with different disasters. Namely the area of burn scars in the \textit{Fire} dataset has both the largest area of effect and also the largest proportion of changed pixels to all not cloudy pixels (reported as positive ratio).

\begin{table}[!htb]
    \caption{Validation dataset details. Each location is captured in 4 time-steps before the event and once after the event (it is however counted only once into the cumulative square km). Note that only the last pair of images is labelled with changed and not changed pixel map - and that the reported percentage only considers this map.}
    \label{tab:dataset-eval-summary}
    \centering
    \begin{tabular}{@{}lcccc@{}}
    \toprule
                  &  Number of &  Cumulative   & Positive rate   \\ 
                        &  locations &  area (km$^2$) & (\%)     \\
                        
    Landslides         & 5       &   108     & 10.48    \\
    Floods             & 4       &  1301     & 6.74     \\
    Hurricanes         & 5       &  1622     & 24.31    \\
    Fires              & 5       &  3485     & 53.79    \\
    \bottomrule
    \end{tabular}
\end{table}

\end{document}